\pdfoutput=1
\documentclass[11pt]{article}

\usepackage{acl}

\usepackage{times}
\usepackage{latexsym}

\usepackage[T1]{fontenc}
\usepackage[utf8]{inputenc}

\usepackage{microtype}

\usepackage{amsmath}    % \text{}
\usepackage{amssymb}    % \mathbb{}
\usepackage{booktabs}   % \toprule \midtule \bottomrule
\usepackage{graphicx}   % {figure}
\usepackage{subcaption}
\usepackage{multirow}

\newcommand{\transformer}{{Transformer}}

\newcommand{\algname}{\textsc{TranSync} }

\title{Efficient Long Sequence Encoding via Synchronization}

\author{Xiangyang Mou \\
  Rensselaer Polytechnic Institute \\
  \texttt{moux4@rpi.edu} \\
  \And
  Mo Yu \\
  WeChat AI, Tencent    \\
  \texttt{moyumyu@tencent.com} \\
  \AND
  Bingsheng Yao \\
  Rensselaer Polytechnic Institute    \\
  \texttt{yaob@rpi.com} \\
  \And
  Lifu Huang \\
  Virginia Tech    \\
  \texttt{lifuh@vt.edu} \\
}

\begin{document}
\maketitle
\begin{abstract}
Pre-trained \transformer\ models have achieved successes in a wide range of NLP tasks, but are inefficient when dealing with long input sequences. Existing studies try to overcome this challenge via segmenting the long sequence followed by hierarchical encoding or post-hoc aggregation. We propose a synchronization mechanism for hierarchical encoding. Our approach first identifies anchor tokens across segments and groups them by their roles in the original input sequence. Then inside \transformer\ layer, anchor embeddings are synchronized within their group via a self-attention module. Our approach is a general framework with sufficient flexibility -- when adapted to a new task, it is easy to be enhanced with the task-specific anchor definitions. Experiments on two representative tasks with different types of long input texts, NarrativeQA summary setting and wild multi-hop reasoning from HotpotQA, demonstrate that our approach is able to improve the global information exchange among segments while maintaining efficiency.
\end{abstract}

%%%%%%%%%%%%%%%%%%%%%%
%%%  Introduction  %%%
%%%%%%%%%%%%%%%%%%%%%%

\section{Introduction}
\transformer-based encoders \cite{vaswani2017attention} have been widely used in natural language processing with successes. The pre-trained language models based on \transformer, such as BERT~\cite{devlin2018bert}, GPT-3~\cite{brown2020language}, T5~\cite{raffel2019exploring} and BART~\cite{lewis2019bart}, further make it a dominating architecture in NLP.

\begin{figure}[!t]
    \centering
    \includegraphics[width=0.5\textwidth]{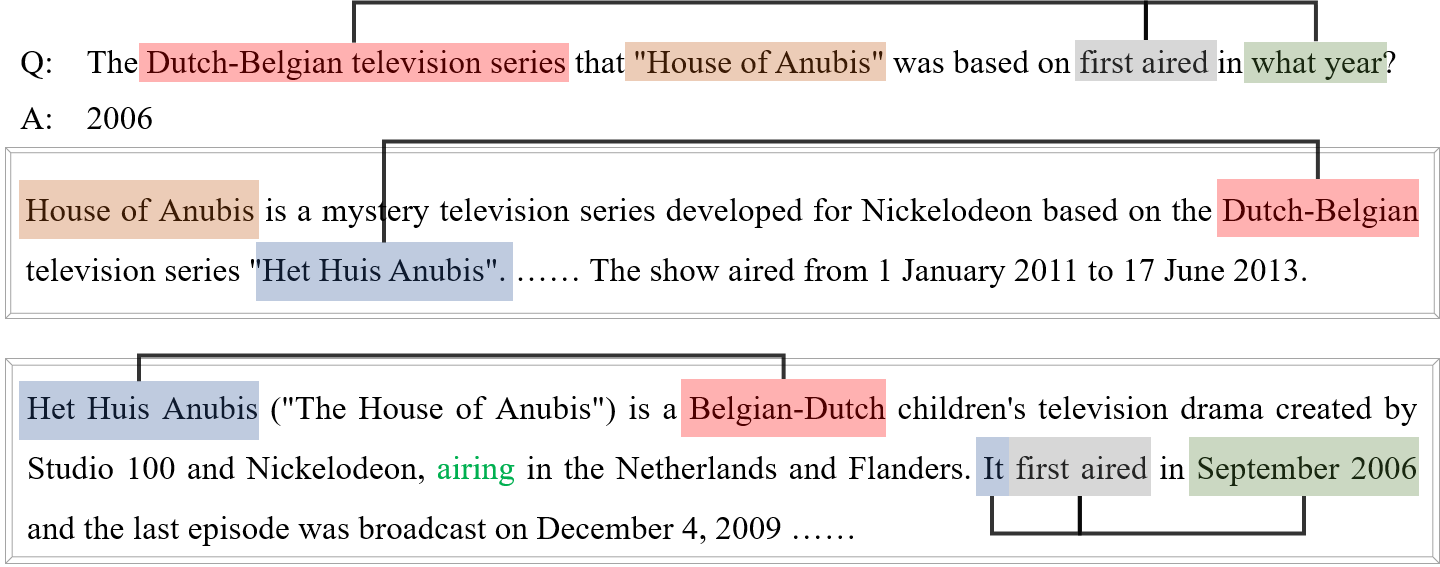}
    \caption{\small{An example from HotpotQA. Different entities are color-coded. The solid lines indicate the correct partial evidence chain toward the true answer. \emph{Dutch-Belgian} is an example of anchors that can pass the partial evidence to the others for collecting full evidence.}}
    \label{fig:intro_example}
    \vspace{-14pt}
\end{figure}

Despite its successes, the \transformer\ models suffer from a major challenge in encoding long sequences. It is due to the fact that the self-attention mechanism used in each \transformer\ layer requires to compute attention for each pair of input words. Such computations lead to $O(l^2)$ complexity in time and space in each \transformer\ layer, where $l$ is the sequence length. This limits \transformer's roles in the increasingly important long sequence encoding for two common scenarios: (1) \emph{encoding of a single long document} with lengths exceeding the input limitation, and (2) \emph{joint-encoding of multiple related documents} for tasks that require synthesizing scattered pieces of evidence, e.g., multi-hop reasoning and multi-document summarization. Figure~\ref{fig:intro_example} gives an example of the necessary information exchange in multi-hop QA. Each paragraph provides a partial clue to solve the task (shown as the connected entities). Intuitively, an effective global encoding should allow the entities (e.g., \emph{Dutch-Belgian}) appearing in multiple paragraphs to share information across all their occurrences. In this way, the embedding of \emph{Dutch-Belgian} in the second paragraph can be aware of partial evidence from the first one and resolve the required information of \emph{House of Anubis}.

To overcome this difficulty, many techniques have been proposed. The existing solutions can be categorized into two classes. The first class is \textbf{hierarchical encoding}. The idea is to either explicitly split the input into multiple short segments for fast encoding of each segment and then exchange information on top of their embeddings following sample-agnostic strategies~\cite{ainslie2020etc,wang2020cluster}; or implicitly constrain the information exchange among tokens with a sparse attention map~\cite{beltagy2020longformer,zaheer2020big}. The essential of these methods is to find efficient ways to pass information among segments and compensate for the loss of important global context across segments. One solution is introducing a pseudo token for each segment and encouraging the pseudo tokens to attend one another during encoding~\cite{ainslie2020etc} for inter-segment interactions.
The second class is \textbf{post-hoc aggregation} in the generative framework, such as Fusion-in-Decoder (FiD, \citealt{izacard2020leveraging}) with BART model. The input is split into segments that are encoded independently by the encoder. Then the decoder casts global attention over all the segments and generates the prediction. This approach allows for only shallow information exchange because the encoding is purely localized; yet is proved empirically very powerful in many NLP tasks.

We propose an orthogonal direction towards an efficient encoding of long sequences. Our method starts with local segments in the post-hoc aggregation approaches and relies on our proposed \textbf{synchronization mechanisms} to swap useful information from the other relevant segments during encoding, so as to maintain global information. Formally, our approach first identifies a set of anchors in the segments and puts them into different groups based on the similarity in their semantic units or the roles they play in the original input sequence. The identified anchors and groups connect different segments logically and naturally.
% In the encoding stage with multiple \transformer\ layers, each layer still works with each short segment independently. After each encoding layer, we synchronize the anchor embedding inside each group, by updating their embedding with the other anchors in the same group via a gating function.
Our synchronization is applied only to the encoding stage where inside each \transformer\ layer of the encoder, we perform an additional embedding update for each anchor using other anchor embeddings in the same group after a normal local encoding. The local encoding and anchor synchronization happen iteratively so that the global information is propagated deeply among segments with anchors as bridges.

Compared to previous hierarchical encoding approaches with fixed communication designs, our approach is more powerful and flexible.
First, our approach provides a finer-grained information exchange mechanism.
Second, our approach is a general framework that reduces the problem of global encoding to synchronization schema design. For any new applications or tasks, it is easy to infuse human prior knowledge to the model by identifying task-specific anchors and anchor grouping.
% Therefore, our approach has better potential to suit the target tasks

% \paragraph{Advantages}
% Flexiblity, a general framework reducing the problem to synchronization scheme design.
% Additionally, the method encourages information propagation in intermediate layers.

We evaluate our approach on two different tasks that require encoding of long sequences: NarrativeQA~\cite{kovcisky2018narrativeqa}, where each input is a single long story summary; and a wild multi-hop QA task adapted from HotpotQA~\cite{yang2018hotpotqa}, where the evidence annotation is assumed unavailable and the input documents are treated more independently from each other. The two settings correspond to the representative examples of the aforementioned long sequence encoding scenarios (1) \& (2).
Results show that our approach significantly improves the performance while remaining efficient.
Moreover, building on top of FiD, the state-of-the-art hierarchical method ETC~\cite{ainslie2020etc} does not bring further improvement as we observe but our approach improves consistently.

%%%%%%%%%%%%%%%%%%%%
%%%    Method    %%%
%%%%%%%%%%%%%%%%%%%%

\section{The Transformer with Synchronization (\algname) Framework}
\label{sec:trans_sync}

In this section, we propose our \algname\  framework which extends \transformer\ layer with an embedding synchronization module attached to the end. Given a long context sequence $C$, we divide it into segments, i.e. $C = [s_1; s_2; ...; s_n]$ where $s_i$ is the $i$-th segment of $C$ and $n$ is the number of segments. A segment can represent a natural sentence or a sequence in a certain length. Together with the question $q$, we re-organize the input to the \transformer\ and form a set of question-prefixed segments $\{s_i^q\}_{i=1}^n$, s.t.
\begin{align}   \label{eq:segment_prefix}
    s_i^q = [q; \text{<SEP>}; s_i]
\end{align}
where $\text{<SEP>}$ is a special token. 
An embedding layer converts the text segments $\{s_i^q\}_{i=1}^n$ into their corresponding question-aware embeddings $\{\mathbf{e}_i^q\}_{i=1}^n$, s.t.
\begin{align}
    \mathbf{e}_i^q = [\mathbf{t}_i^1; \mathbf{t}_i^2; ...; \mathbf{t}_i^{l_i}] \in \mathbb{R}^{l_i \times d}
\end{align}
where $l_i$ is the length of $s_i^q$'s token sequence, $d$ is the dimension of the feature vector and $\mathbf{t}_i^j \in \mathbb{R}^d$ is the embedding for the $j$-th token in the $i$-th segment.

For genericity, our synchronization is performed between the target anchor and the incoming anchors, following the idea of message passing. The values of the target anchor embedding $\mathbf{a}_t$ are updated with the weighted sum of the incoming anchor embeddings and itself, i.e.,
% \begin{align}
%     \overline{\mathbf{a}_{in}} = \sum_{k} \alpha_k \mathbf{a}_{in}^k    \label{eq:sync_target}  \\
%     (\mathbf{a}_t)' = (1 - \beta_t) \cdot \mathbf{a}_t + \beta_t \cdot \overline{\mathbf{a}_{in}} 
% \end{align}
\begin{align}
    \mathbf{a}_t' = \sum_{k} \alpha_k \mathbf{a}^k, \quad s.t. \sum_k \alpha_k = 1   \label{eq:sync_target}  
\end{align}
where $\{\mathbf{a}^k\}$ are the embedding spans\footnote{Some words may correspond to multiple tokens due to the byte pair encoding (BPE) algorithm.} of the same length within the same anchor group; $\alpha_k$ is the normalized weight. In this work, for each anchor group, we form a new sequence from the selected anchor embeddings, i.e., $[ \mathbf{a}^1; \mathbf{a}^2; ...; \mathbf{a}^k]$, and use a self-attention module to compute the weights and update the embedding values.

Our \algname\ framework is embedded in \transformer\ layer. The synchronization is performed between the local self-attention and normalization steps to achieve deep information exchanging. At the end of the last \transformer\ layer, the synchronized segment embeddings are fused into one by concatenating one another as follows:
\begin{align}   \label{eq:segment_fusion}
    [\mathbf{e}_1^q; \mathbf{e}_2^q; ...; \mathbf{e}_n^q] \in \mathbb{R}^{\sum_1^n l_i \times d}
\end{align}
The flexibility of our \algname\ framework is granted by the manifold strategies of identifying anchors in the segments and the heterogeneous message passing directions. The schema will be detailed in Section~\ref{sec:setting}.

\section{Evaluating Tasks}
\label{sec:setting}

In this section, we introduce two experiments performed to verify the feasibility and flexibility of our \algname\ framework. 

\subsection{NarrativeQA}
\label{ssec:narqa}

\paragraph{Task Description}
NarrativeQA dataset has a collection of 783 books and 789 movie scripts. Each book and script is annotated with a long summary and 30 question-answer pairs on average. NarrativeQA provides two different settings, the summary setting and the full-story setting. In this work, we follow the summary setting by answering questions from the summaries, and formulate it as a generative QA task due to the free-form annotated answers.
NarrativeQA is a representative example of the first type of long sequence encoding scenarios:
a single document with length exceeding the input limitation.

\paragraph{Synchronization Schema}
We split each summary into natural sentences as the segments $\{s_i\}$ and prefix them with the question $q$ following Section~\ref{sec:trans_sync}. This breakup of the continuous summary sentences drops the global context across segments during encoding. As a compensation, we apply a segment-level synchronization, which takes the preceding question sequence as the anchor. Practically, we simply use the special token <SEP> that connects to the question in each segment as the representatives, which significantly reduce the synchronization cost. The segment-level synchronization happens only among the closest neighbouring segments, inspired by their natural order in the summary text. Intuitively, it provides each segment compressed contextual information from its neighbors; and makes the question embedding be aware of its matched contents across multiple segments. Therefore, we expect it can better deal with questions that require multiple sentences to answer.

%%%%%%%  VERSION 1  %%%%%%%
% \paragraph{Synchronization Schema}
% We split each summary into natural sentences as the segments $\{s_i\}$ and prefix them with the question $q$ following Section~\ref{sec:trans_sync}. This breakup of the continuous summary sentences drops the global context across segments during encoding. As a compensation, we apply a segment-level synchronization, which takes the preceding question sequence as the anchor. The segment-level synchronization happens only among the closest neighbour, i.e.,
% \begin{align}
%     \overline{\mathbf{a}_{in}} =
%     \begin{cases}
%         \mathbf{a}_2        ,   & i = 1    \\
%         \frac{1}{2} \sum_{k \in \{i-1, i+1\}} \mathbf{a}_k, & 1 < i < n \\
%         \mathbf{a}_{n-1}    ,   & i = n   \\
%     \end{cases}.
%     \label{eq:schema1}
% \end{align}
% This synchronization schema allows information exchange between every pair of neighbor sentences.
% Intuitively, it provides each segment compressed contextual information from its neighbors; and makes the question embedding be aware of its matched contents from multiple segments.
% Therefore, we expect it can better deal with questions that require multiple sentences to answer.

\subsection{Wild Multi-hop Reasoning}
\label{ssec:hotpot}

\paragraph{Task Description}
We construct a wild multi-hop reasoning task from the HotpotQA dataset which provided two evidence documents and eight distractor documents for each question. We adopt the realistic assumption with no evidence annotation provided, to investigate the models' ability to sort out the reasoning chains from multiple documents. We designed two settings on the HotpotQA dataset intending to verify the effect of various context lengths on different models. The \textbf{MultiHop-10} uses all the 8 distractors in the dataset, the concatenation of the documents is thus beyond the length limit of BART. The \textbf{MultiHop-6} uses only 4, which is on average within BART's limit. 

% The dataset offers additional sentence-level evidence annotations which is assumed to be unavailable in the wild setting. 
With the wild multi-hop reasoning, we hope to justify if our \algname\ can effectively pass important messages across segments. For consistency, we also formulate it as a generative QA task with the goal of predicting a free-form or YES/NO answer given the question and the context.

\paragraph{Synchronization Schema}
We split each concatenated document into segments in similar lengths containing various numbers of natural sentences. We have two ways of synchronization according to the task's unique properties.
Firstly, a similar segment-level synchronization schema is applied. However, due to the different segment splitting strategies, there is no continuation guaranteed between the neighboring segments. Therefore, we synchronize across all segments rather than only among the neighbors.
Secondly, we take the titles of the original documents as word-level anchors. For simplicity, the titles are added to the input\footnote{The title words already appear in the document, hence adding them to the input does not introduce new information and is regarded as a fair comparison.}, immediately following the question sequence. Similarly, we perform synchronization among all the title-associated special tokens to cut down computational costs. Due to the multi-hop nature of the samples, we expect the token-level synchronization to help build latent connections among the evidence.

%%%%%%%  VERSION 1  %%%%%%%
% \subsection{Evidence Ordering Task from HotpotQA}
% \label{ssec:hotpot}

% \paragraph{Task Description}
% \citet{wang2019dream} proposes an evidence ordering task from the HotpotQA dataset, which provides the two evidence paragraphs of each question in a random order and asks a QA model to find the answer. This ordering task corresponds to the second representative type of long sequence encoding scenarios: multiple related documents that require joint-encoding. With the specific multi-hop reasoning challenge, we hope to verify if our \algname\ can effectively pass important messages across segments. For consistency, we also formulate it as a generative QA task with the goal of predicting a free-form or YES/NO answer given the question and the context.

% \paragraph{Synchronization Schema}
% We have two ways of synchronization according to the task's unique properties.
% Firstly, we directly use the sentence splits provided by HotpotQA and prefix them with the question sentence. A similar segment-level synchronization schema is applied, but only to the sentences from the same paragraph.
% Secondly, we take the titles of the documents as the word-level anchors and synchronize them with the same words and phrases in the segments. Due to the multi-hop nature of the samples, we expect the extra synchronization among the entities to provide clues of the evidence order for the neural models.

\begin{table}[h]
        \small
        \centering
        \begin{tabular}{p{12mm}ccccc} 
            \toprule
            \multirow{2}{*}{\bf System} & \bf NarQA & \multicolumn{2}{c}{\bf MultiHop-10}  & \multicolumn{2}{c}{\bf MultiHop-6} \\
            & \bf Rouge-L & \bf EM & \bf F1 & \bf EM & \bf F1\\
            \midrule
            % SOTA        & 58.76 &  -    &  -    &   -   &   -     \\
            BART        & 64.78 & 41.63 & 54.85 & 55.62 & 69.96   \\
            FiD         & 66.57 & 55.65 & 69.35 & 57.42 & 71.30   \\
            FiD+ETC     & 65.89 & 55.46 & 69.31 & 57.52 & 71.66   \\
            \algname   & \bf 67.58 & \bf 56.49 & \bf 70.32 & \bf 58.30 & \bf 72.61   \\
            \bottomrule
      \end{tabular}
      \caption{Overall results on the NarrativeQA and the two multi-hop setting tasks (\%). }
    %   The SOTA result on NarrativeQA dataset is based on~\cite{mou2020frustratingly}.}
      \label{tab:results}
    \end{table}

\vspace{-2mm}
\section{Experiments}
\label{sec:exp}

\paragraph{Baseline}
Our backbone model is the pre-trained BART-{large} model\footnote{Implementation from \url{https://huggingface.co/}}. We compare with three baselines:
(1) the original \textbf{BART}, which directly takes the concatenation of the question and the raw sequence without splitting. The sequence is truncated with a maximum of 1,024 tokens.
% Since this method suffers from the inefficiency when encoding long sequences, we truncate the input length with 1,024 tokens.
(2) \textbf{FiD}~\cite{izacard2020leveraging}, the state-of-the-art hierarchical encoding algorithms for generative \transformer\ models.
(3) \textbf{FiD+ETC}, a FiD variant enhanced by our implementation of ETC~\cite{ainslie2020etc} in the encoder.

% Our second baseline is ETC~\cite{ainslie2020etc}, the state-of-the-art hierarchical encoding algorithm for \transformer.

\paragraph{Metrics}
Because of the generative nature of the NarrativeQA task, following previous works~\cite{kovcisky2018narrativeqa,tay2019simple,mou2020frustratingly}, we evaluate the QA performance with Rouge-L~\cite{lin2004rougeL}.\footnote{We use an open-source evaluation library~\cite{sharma2017nlgeval}: \url{https://github.com/Maluuba/nlg-eval}.}
On HotpotQA dataset, the Exact Match (EM) and F1 scores\footnote{The squad/evaluate-v1.1.py script is used.} are reported that are commonly used in open-domain QA evaluation. Both hypothesis and reference are lowercased with the punctuation removed before evaluation.

% \subsection{Results}

\paragraph{Overall Results}
Table~\ref{tab:results} shows the overall results on all three tasks.
Our proposed \algname achieves the best results on both NarrativeQA and our new wild multi-hop QA tasks.

To our surprise, splitting the long sentences into question-aware segments alone (FiD) gives strong results against the BART baseline. This indicates the post-hoc aggregation of local embeddings can handle a significant portion of testing cases, reflecting the absence of global reasoning in many existing datasets. Our synchronization mechanism compensates for the loss of global context resulting from the sequence splitting and brings a consistent $~1\%$ improvement over FiD across all three tasks. ETC does not provide a further improvement over FiD as our approach does. This empirically shows that ETC's synchronization mechanism does not provide complementary global information to the post-hoc aggregation approach.

% \paragraph{NarrativeQA}
% Our proposed \algname achieved the new state-of-the-art (SOTA) performance on NarrativeQA dataset with $4.3\%$ improvement over the BART baseline on Rouge-L.
% and $15.0\%$ over the previous SOTA, elaborating the contributions of our proposed framework and synchronization schema. 
% To our surprise, splitting the long sentences into question-aware segments alone (FiD) contrarily brought a $1.79$-points rise in Rouge-L, considering the absence of global encoding. We reckoned that it facilitated the model to better learn the local context from short segments and overcame the encoding inefficiency for long sequences. 
% Our synchronization among the questions compensated the loss of global context resulted from the sequence splitting and brought a further $1.01\%$ improvement. 
% Our additional ablation study showed that the neighbouring syncing schema earned slightly more gains by $0.3$ point in Rouge-L score than all-segment syncing schema, which probably attributed to less data disturbance from avoiding unnecessary synchronization. It also partially explained why ``FiD + ETC'' has a lower score than ``FiD'' on NarrativeQA dataset.

% \paragraph{Multi-hop Reasoning}
% to eliminate the impact on BART models input truncation.
% Table~\ref{tab:results} shows that our \algname\ brought consistent improvements over all the baselines in both settings, demonstrating the benefit of exchanging information across segments for reasoning tasks. The advantage of \algname\ over ETC lies at token-level synchronization rather than coarse-grain interaction.

Finally, aside from Table~\ref{tab:results}, we also experiment with different segment lengths and find that the split context length should be at least 2 times longer than the prefixing question for effective encoding; otherwise, the question would dominate in the segment and it would lead to a significant drop in performance. Together with the observations that FiD with short segments outperforms BART with long sequences in both settings, we conclude that the splitting length is a hyper-parameter worth tuning.

% The different observations on two datasets attribute to the low length ratio between the context and question sequences. The relatively long question sequence in HotpotQA dominates in the naturally-splitted question-aware segments and makes the local encoding significantly inefficient. Our ablation study shows that the model also benefits from our synchronization mechanism with nearly 1 point increase in EM. However, the gain from global encoding is not significant enough to make up for the performance drop due to the inefficient local encoding.

% which drives us to take the advantage of the flexibility of our synchronization mechanism and come up with an improved schema for future work.

% It reveals that our methods is more suitable 

\paragraph{Efficiency}
Table~\ref{tab:efficiency} provides an analysis of the efficiency of our \algname\ framework.
The complexity comparison shows that the \algname\ is more memory efficient than the BART baseline in theory and becomes more superior when $l_q \ll \frac{l_c}{n}$.
We also compare the runtime speed empirically by measuring the average time used for encoding per token. Though the synchronization introduces extra complexities to the encoding procedure, our experiments on the NarraitveQA dataset verify that the overall speed of our methods remains doubled to the BART baseline.

% The bidirectional nature of the \transformer\ encoder leaves the complexity of BART to be $\mathcal{O}(l^2)$ during encoding, where $l$ is the length of the input sequence. Table.~\ref{tab:efficiency} shows that our \algname\ can be much more memory- and time-efficient in theory when $l_q \ll \frac{l_c}{n}$. Though the synchronization introduces extra complexities to the encoding procedure, our empirical experiments in inference phase under the settings of Table.~\ref{tab:narrativeqa} verify that the overall efficiency of \algname\ remains more than doubled. 

\begin{table}[h]
    \small
    \centering
    \begin{tabular}{lcc} 
        \toprule
        \bf System      & \bf $\mathcal{O}(f)$                  & \bf Time/Token  \\
        \midrule
        BART Baseline   & $(l_q + l_c)^2$                       &   292  $\mu$s       \\
        \algname\       & $(l_q + \frac{l_c}{n})^2 \cdot n$     &   134  $\mu$s       \\
        \bottomrule
    \end{tabular}
    \caption{Efficiency comparison. $l_q$ and $l_c$ are the length of the question sequence and the context sequence; $n$ is the number of split segments. The encoding time per token is averaged over 100 QA samples.}
    \label{tab:efficiency}
\end{table}

%%%%%%%%%%%%%%%%%%%%
%%%  Conclusion  %%%
%%%%%%%%%%%%%%%%%%%%

\vspace{-3mm}
\section{Conclusion}
In this work, we propose \algname\ framework with flexible synchronization mechanisms for encoding long sequences. We demonstrate the feasibility of our method in reasoning tasks with long context, and also show its high adaptability to different scenarios. We consider our work to be valuable as an easy solution to address the long context issue in QA, and to be potentially applicable to other long sequence modeling tasks.

\bibliography{anthology,custom}
\bibliographystyle{acl_natbib}

\appendix

\end{document}